\documentclass[conference]{IEEEtran}

\IEEEoverridecommandlockouts
\usepackage{multirow}
\usepackage{CJKutf8}
\usepackage{cite}
\usepackage{amsmath,amssymb,amsfonts}
\newtheorem{definition}{Definition}
\usepackage[ruled, linesnumbered, vlined]{algorithm2e}
\usepackage{graphicx}
\usepackage{wrapfig}
\usepackage{subcaption}
\usepackage{svg}
\usepackage{textcomp}
\usepackage{xcolor}
\usepackage{booktabs}
\def\BibTeX{{\rm B\kern-.05em{\sc i\kern-.025em b}\kern-.08em
    T\kern-.1667em\lower.7ex\hbox{E}\kern-.125emX}}
\begin{document}
\bibliographystyle{unsrt}

\title{Business Process Text Sketch Automation Generation Using Large Language Model}

\author{

\IEEEauthorblockN{1\textsuperscript{st} Rui Zhu}
\IEEEauthorblockA{\textit{School of Software} \\
\textit{Yunnan University}\\
Kunming, China \\
rzhu@ynu.edu.cn}

\\

\IEEEauthorblockN{4\textsuperscript{th} Honghao Xiao}
\IEEEauthorblockA{\textit{School of Software} \\
\textit{Yunnan University}\\
Kunming, China \\
709964695@qq.com}

\and

\IEEEauthorblockN{2\textsuperscript{nd} Quanzhou Hu}
\IEEEauthorblockA{\textit{School of Software} \\
\textit{Yunnan University}\\
Kunming, China \\
hqz3232@qq.com}

\\

\IEEEauthorblockN{5\textsuperscript{th} Chaogang Wang}
\IEEEauthorblockA{\textit{School of Software} \\
\textit{Yunnan University}\\
Kunming, China \\
wangchaogang0220@163.com}

\and

\IEEEauthorblockN{3\textsuperscript{rd} WenXin Li}
\IEEEauthorblockA{\textit{School of Software} \\
\textit{Yunnan University)}\\
Kunming, China \\
1614145921@126.com}

\\

\IEEEauthorblockN{6\textsuperscript{th} Zixin Zhou}
\IEEEauthorblockA{\textit{School of Software} \\
\textit{Yunnan University}\\
Kunming, China \\
2924845590@qq.com}

}

\maketitle

\begin{abstract}
Business Process Management (BPM) is gaining increasing attention as it has the potential to cut costs while boosting output and quality. Business process document generation is a crucial stage in BPM. However, due to a shortage of datasets, data-driven deep learning techniques struggle to deliver the expected results. We propose an approach to transform Conditional Process Trees (CPTs) into Business Process Text Sketches (BPTSs) using Large Language Models (LLMs). The traditional prompting approach (Few-shot In-Context Learning) tries to get the correct answer in one go, and it can find the pattern of transforming simple CPTs into BPTSs, but for close-domain and CPTs with complex hierarchy, the traditional prompts perform weakly and with low correctness. We suggest using this technique to break down a difficult CPT into a number of basic CPTs and then solve each one in turn, drawing inspiration from the divide-and-conquer strategy. We chose 100 process trees with depths ranging from 2 to 5 at random, as well as CPTs with many nodes, many degrees of selection, and cyclic nesting. Experiments show that our method can achieve a correct rate of 93.42\%, which is 45.17\% better than traditional prompting methods. Our proposed method provides a solution for business process document generation in the absence of datasets, and secondly, it becomes potentially possible to provide a large number of datasets for the process model extraction (PME) domain.
\end{abstract}

\begin{IEEEkeywords}
Business Process Management, Business process document generation, Conditional Process Tree, Business Process Text Sketch, divide-and-conquer
\end{IEEEkeywords}

\section{Introduction}
\begin{figure*}[t]
\centerline{\includegraphics[scale=0.5]{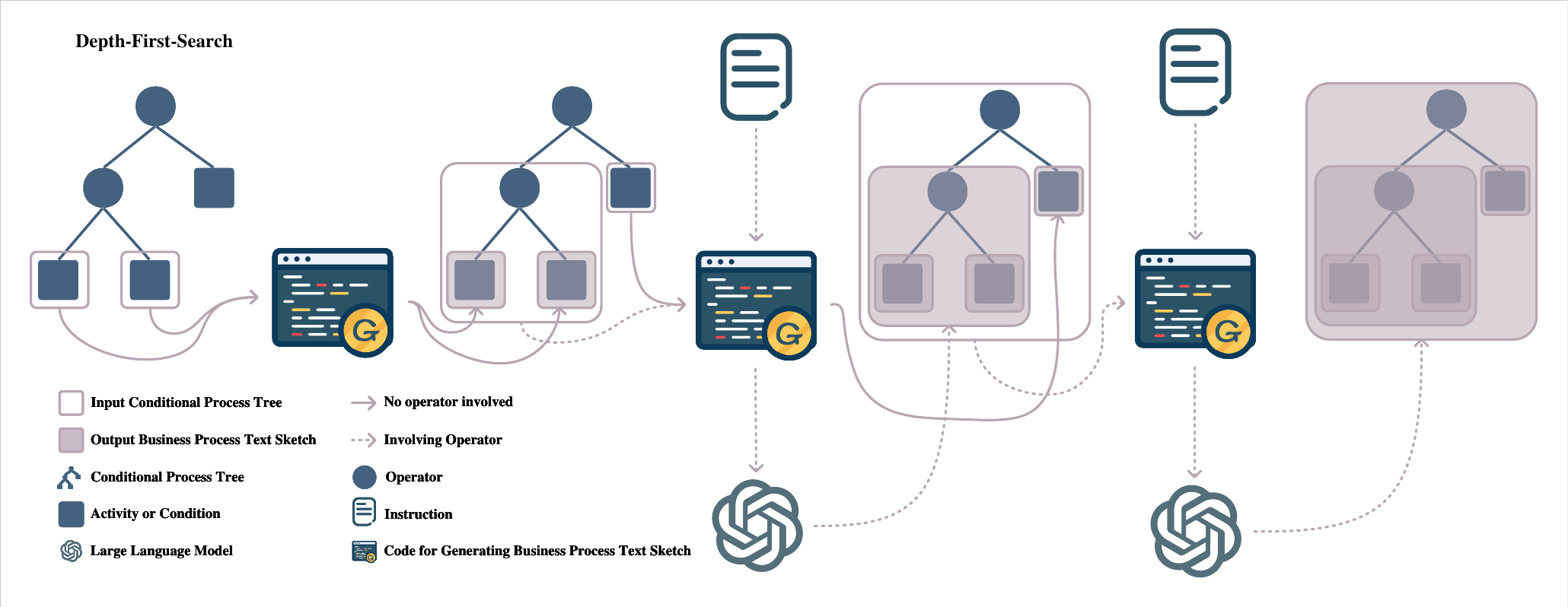}}
\caption{Overall framework. Through a deep cross-track way through the CPT, each processing of a single layer of the CPT is divided into two ways. a. No operator. By using code or rules, BPTS can be generated when processing activities or conditions. b. Involving operators. When operators are involved, it is necessary to consider how to smoothly connect multiple BPTSs into one BPTS, and we draw on the knowledge of LLMs to help us accomplish the task.}
\label{fig:framework_new}
\end{figure*}

With the potential to reduce costs, improve productivity, and achieve higher levels of quality, Business Process Management (BPM) has been attracting more and more attention recently \cite{han2020bps}. As an important part of BPM, business process document generation can help companies clarify and standardize business processes and organize and sort out business. On the other hand, business process documentation can be used as an important tool for enterprise communication and collaboration. However, due to the lack of datasets in the field, it is difficult to deliver the performance that deep learning deserves. A closely related domain is Process Model Extraction (PME), which aims to extract business process models from business process description texts using Natural Language Processing (NLP) techniques. Annotated data in this domain is similarly scarce, and data for much of this domain is often not publicly available. To make matters worse, most of the available data is somehow extracted from the dataset originally proposed by Friedrich \cite{friedrich2011process} and can hardly be considered as a reference dataset in practice. The only exception is the dataset proposed by Qian \cite{qian2020approach}, but this dataset is only annotated with words and sentences and does not have a complete process model annotation \cite{bellan2021process}. Consequently, it is also impossible to use the dataset from the PME domain.

In recent years, Large Language Models (LLMs) \cite{brown2020language,openai2023gpt4,bubeck2023sparks} have achieved surprising performance on many NLP tasks. To activate the capabilities of LLMs, prompt learning \cite{schick2020exploiting} and In-Context learning \cite{min2022rethinking}, and prompt engineering\footnote{https://github.com/dair-ai/Prompt-Engineering-Guide} are beginning to emerge. Prompt Learning is mainly used in low-resource scenarios, such as Zero-shot learning and Few-shot learning. According to this, we propose a method of using LLMs to automatically generate Business Process Text Sketchs (BPTSs) to alleviate the awkward situation of lacking datasets in the field of business process document generation. The embodiment of BPTSs is natural language, which ignores the semantics of activities and conditions while maintaining the time links between activities in business process descriptions. Theoretically, it can be instantiated as a business process description text for any domain. So when the Conditional Process Tree (CPT) is instantiated, the generated BPTS will also be the instantiated business process description text, thus achieving the purpose of business process document generation. It is worth mentioning that although BPTS ignores the semantic information of activities and conditions, it retains the framework for describing the entire process, including hierarchical and timing relationships. Perhaps what the PME field needs to learn is also about hierarchical and temporal relationships. We can generate a large number of CPT-BPTS pairs using our rules and approach, thus providing a potential possibility to expand the PME domain dataset.

However, traditional prompting methods (Few-shot In-Context Learning) have difficulty in performing extremely well on close-domain and transformation problems with complex hierarchical structures. In order for LLMs to learn the rules for the close-domain problem, we need to give them a lot of domain knowledge and examples, which makes it challenging to manage the prompt template. Currently, there is no reliable method to find the best prompt, and since we use tree-structured input, the CPTs need to be serialized when fed into LLMs using traditional prompting methods, which results in the structural and node information of the CPTs being compressed into one dimension. LLMs not only need to parse the hierarchical structure of CPTs but also have to learn how to translate the hierarchy into the corresponding natural language representation, which increases their burden. We offer an automatic BPTS generation approach to break down a large CPT into numerous basic CPTs and then solve each one separately, drawing inspiration from the divide-and-conquer strategy. For the simple CPTs, we no longer consider them as inference transformation tasks but language rewriting tasks that LLMs are better at. Our main contributions are as follows:
\begin{itemize}
\item For the first time, we suggest using LLMs to create business process documentation in order to address the dearth of datasets in this field.
\item In order to increase the accuracy of CPT to BPTS conversion, we suggest using the divide-and-conquer strategy to break down a complex CPT into simpler CPTs and then solve each one in turn.
\end{itemize}

The related work is introduced in the second chapter of this paper, and the definitions of the CPT, BPTS, and language rewriting tasks are introduced in the third chapter. The method portion is covered in the fourth chapter, the experimental section is covered in the fifth chapter, this paper's limits are covered in the sixth chapter, and the conclusion and discussion are covered in the sixth chapter.

\section{Related Work}
\subsection{Process Model Extraction}
There is currently some work accumulated in the Process Model Extraction (PME) field, but many of these datasets are not publicly available. Some publicly available datasets have only a few dozen business process description texts, while others only annotate sentences and words without a complete process model annotation. It is impossible to apply the PME domain dataset to the field of business process document generation. Han et al.\cite{han2020bps} proposed the use of Ordered Neuron LSTM (ON-LSTM) to learn language models and their potential semantic hierarchy from Process Definition Documents (PDDs). The dataset they used consists of 210 PDD document graph pairs, which are not publicly available. Qian et al.\cite{qian2020approach} proposed the Multi Grained Text Classifier (MGTC) to classify sentences in process texts. They conducted experimental evaluations on two publicly available datasets, each containing 2636 and 2172 sentences, constructed from cooking recipes and maintenance manuals, which only labeled the words of some sentences. Zhu et al. \cite{lcl} improved the Named Entity Recognition (NER) method to identify the active entities in the business process description text and then used ON-LSTM in an unsupervised discovery process to describe the potential hierarchy between the active entities contained in the document. Zhu et al. \cite{zhu2023tag} found that there are some signal entities such as ``if'', ``after'', etc. in the business process description text. They first obtain the signal entities, conditional entities, and activity entities in the text through sequence annotation, which are called semantic roles. Then, based on predefined rules, a Graph Neural Network (GNN) is constructed using semantic role sequences to predict the temporal relationships of activities in business process documents through multi-layer semantic fusion. They provided some unmarked business process description text. Bellan et al. \cite{bellan2022extracting} utilized large-scale Pre-Training Models (PTMs) and In-Context Learning to directly extract activities, participants, and the execution relationships between participants and the activities they perform from business process description texts. Use a small amount of data.
\subsection{Easy to hard}
The divide-and-conquer strategy, which divides difficult problems into manageable ones and solves them sequentially to produce effective results, serves as the inspiration for some works. Zhou et al. \cite{zhou2022least} proposed a Least-to-Most prompt method, which is divided into Decomposition and Subproblem Solving. For mathematical reasoning problems, they first decompose the problem through LLMs, and then input the decomposed problems into LLMs in turn to obtain the answers. Gao et al. \cite{gao2023easy} proposed Easy-to-Hard Learning (E2H) to solve the problem of Information extraction. They divided Information extraction tasks into easy stages, hard stages, and main stages. They first trained on the task of the easy stage, then on the task of the hard stage, and finally on the task of the main stage. The goal of the easy stage is to enable the model to learn basic skills to help solve major tasks. The goal of the Hard stage is to construct training samples that are more difficult than the original training samples used to train the model. Main stage training is the main task.

\section{Preliminaries}
\begin{definition}
\label{cpt}
Conditional Process Tree. Let A be a finite set of activities, with $\tau\notin A$ representing a silent activity. Let $C$ be a finite set of conditions, $c \in C$ is an arbitrary condition. $\oplus = \{\to,\times,\wedge,\propto\}$ is the set of Conditional Process Tree operators. A Conditional Process Tree is recursively defined as follows:
\begin{itemize}
\item if $a\in A\cup \{\tau \}$, then $Q=a$ is a Conditional Process Tree,
\item if $c\in C$, then $Q=a$ is a Conditional Process Tree and
\item if $Q_1,Q_2,…,Q_n$ are Conditional Process Trees where n$ \ge $1, and $\oplus\in\{\to,\vee\}$, then $Q = \oplus(Q_1,Q_2,…,Q_n)$ is a Conditional Process Tree and
\item if $Q_x,Q_y$ are Conditional Process Trees and $c\in C$ is a condition, then $Q=\times\_c(Q_x,Q_y)$ is a Conditional Process Tree and
\item if $Q_x$ is a Conditional Process Tree and $c\in C$ is a condition, then $Q=\propto(c,Q_x)$ is a Conditional Process Tree.
\end{itemize}
\end{definition}

\begin{figure}[htbp]
\centerline{\includegraphics[scale=0.8]{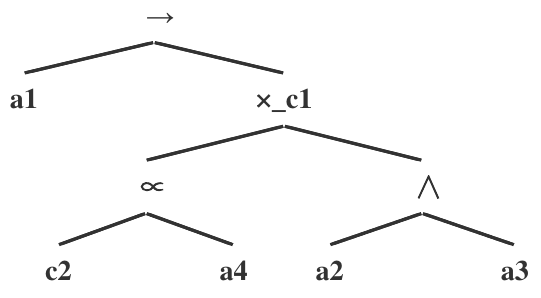}}
\caption{An example of a CPT}
\label{fig:cpt}
\end{figure}

According to our definition, the first child node in the CPT is the one on the left, increasing in number from left to right. An example of a CPT is shown in Figure \ref{fig:cpt}. The sequence node ``$\to$'' indicates that its children are executed in order from left to right. The exclusive node ``$\times\_c$'' indicates that the first child node is executed when the condition ``$c$'' is satisfied, otherwise, the second child node is executed. The loop node ``$\propto$'' indicates that the loop determines whether the condition of its first child node is satisfied, executes the second child node if it is satisfied, and quits the loop if it is not. The parallel node ``$\wedge$'' means that its children are executed simultaneously. In the figure, $a1$, $a2$, $a3$ and $a4$ are codes representing abstract activities. $c1$ and $c2$ are also codes representing abstract conditions. This CPT can be serialized as ``$\to(a1,\times\_c1(\propto(c2,a4),\wedge(a2,a3)))$''. Its corresponding BPTS may be ``Executing activity a1. If condition c1 is met, the loop judges whether condition c2 is met. If it is met, activity a4 is executed, and once condition c2 is not met, the loop ends. If condition c1 is not met, both activities a2 and a3 are executed at the same time.

\begin{definition}
\label{bpts}
Business Process Text Sketch. It is a framework for describing business processes that maintains the timing relationships between activities and ignores the semantic information of activities and conditions.
\end{definition}

\begin{definition}
\label{lrt}
Language Rewriting Task. It refers to using the knowledge of LLMs to convert input into a natural language that means the same and is consistent with human descriptive habits.
\end{definition}

\section{Methodology}
For simple Conditional Process Trees (CPTs), traditional prompts and Large Language Models (LLMs) can solve the problem, but for complex CPTs LLMs often cannot find patterns. The definition of CPT is a recursive reference definition \ref{cpt}, where a subtree of CPT is also CPT. When processing a CPT, its subtree can be processed in the same way before processing itself. In CPT, the parent node of subtrees is an operator that represents how subtrees are combined. Hence, when subtrees are solved, the parent tree can be simply merged with and solved by the subtrees, and the independent subtrees. To sum up, the conversion of CPT can be solved through the divide-and-conquer algorithm method.

The overall framework of our method is shown in Figure \ref{fig:framework_new}. Firstly, we use a deep traversal method to traverse the entire CPT tree. The deep traversal will start at the leaf node, first traverse its child nodes, and then return to itself. This feature is very suitable for our method, where we first build a simple subtree and use the answers obtained to build a more complex parent tree. There are two ways to generate BPTS in the figure. The first is that a leaf node describes an action or a circumstance that does not require LLMs to address it when it is encountered. It is already the smallest unit described in the entire process. The second type is encountered when encountering non-leaf nodes, which are often operators that represent the way in which their sub-activities (resolved) are combined (sequential, exclusive, parallel, loop). Therefore, we use LLMs to handle the operator and its sub-activities. So the input of the prompt needs to be dynamically constructed through code, but the instruction of the prompt remains unchanged.

\subsection{Divide And Conquer}
The LLM is a sequence-to-sequence model, so when we want to input CPT, we need to serialize it, which will result in the structure information and node information of CPT being compressed into a sequence. This makes it necessary for LLMs to parse the hierarchical structure of CPT and accurately combine the hierarchical structure and node information to transform CPT into a corresponding Business Process Text Sketch (BPTS) at once, increasing the difficulty of transformation. We build BPTS from the underlying simple CPT through deep traversal, drawing inspiration from the divide-and-conquer strategy, and then use the built BPTS as a known outcome to build shallower BPTS. Complex CPT transformation problems will be broken down into a number of simpler ones using our approach. In the algorithm, we only process one layer of simple CPT each time. Please refer to Algorithm \ref{rp} for specific details.
\begin{algorithm}[!h]
    \SetKwData{subBPTS}{subBPTS} 
    \SetKwData{BPTS}{BPTS} 
    \SetKwFunction{RecursiveGen}{RecursiveGen}
    \SetKwFunction{append}{append} 
    \SetKwFunction{constructPrompt}{constructPrompt} 
    \SetKwData{prompt}{prompt} 
    \SetKwData{pList}{pList}
    \SetKwData{child}{child}
    \SetKwFunction{ChatGPT}{ChatGPT} 
    \SetKwInOut{Input}{input}
    \SetKwInOut{Output}{output} 
    
    \Input{Instruction $I$ and Conditional Process Tree $NODE$} 
    \Output{Business Process Text Sketch}
    \BlankLine 
    \emph{This is a recursive algorithm called \RecursiveGen}\;
    \eIf{The current node $NODE$ is a non-leaf node}
    {
        \For{\child in $NODE$.$children$}
        {
            \tcp{Recursively call yourself}
            \subBPTS $\leftarrow$ \RecursiveGen{$I$,\child}\;
            \tcp{Storage sub Process Sketch}
            \pList.\append{\subBPTS}\;
        }
        \prompt $\leftarrow$ \constructPrompt{$I$,\pList,$NODE.elem$}\;
        \tcp{Prompt Large Language Models}
        \BPTS $\leftarrow$ \ChatGPT{\prompt}\;
        \textbf{return} \BPTS\;
    }
    {
        \eIf{$NODE$.$type$ == $activity$}
        {
            \BPTS $\leftarrow$ $'execute$ $activity '$ + $NODE.elem$\;
        }
        {
            \BPTS $\leftarrow$ $NODE.elem$\;
        }
        \textbf{return} \BPTS\;
    }
    \caption{Business Process Text Sketch generation algorithm}
    \label{rp} 
\end{algorithm}

\subsection{Task Conversion}
\begin{figure}[!h]
\centerline{\includegraphics[scale=0.5]{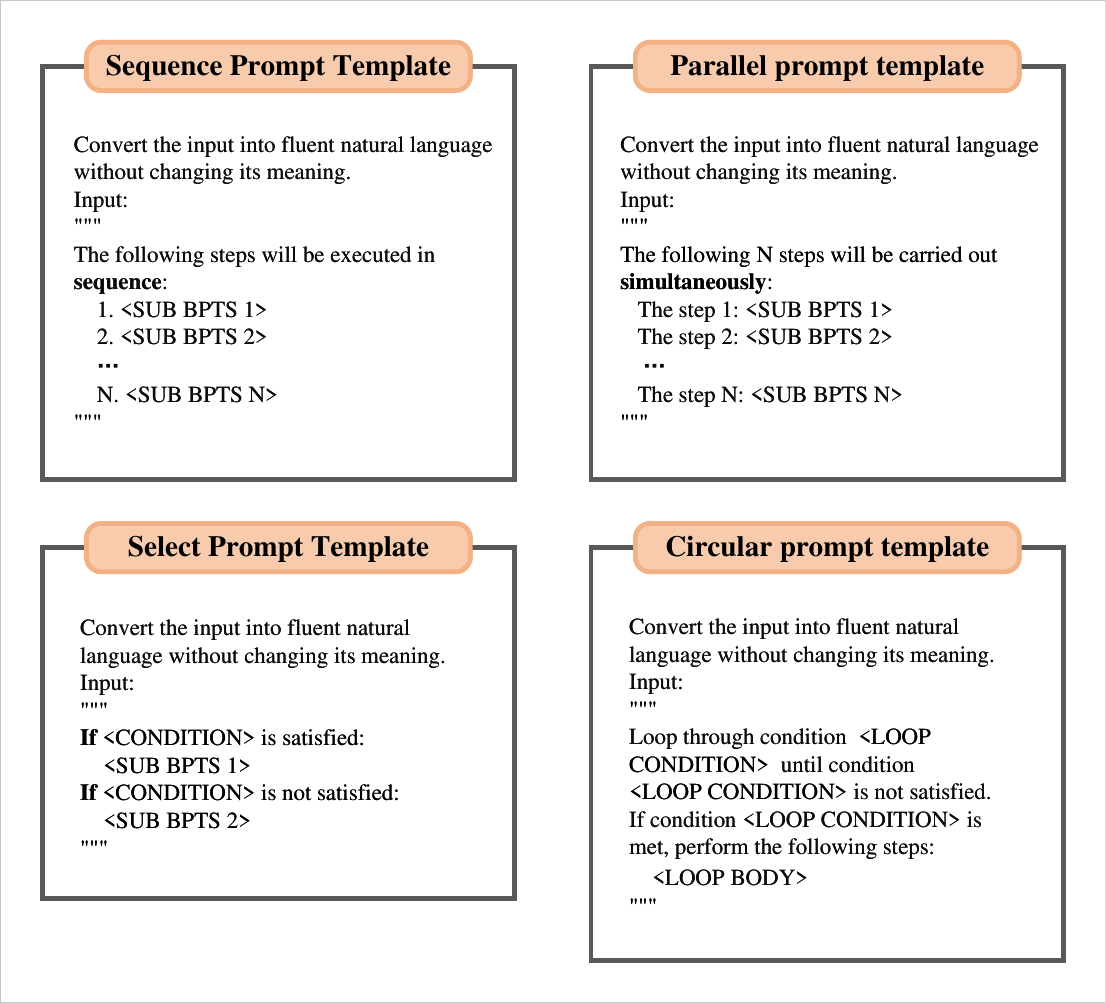}}
\caption{Prompt templates corresponding to the four operators}
\label{fig:prompt_template}
\end{figure}

To reduce the decline in performance caused by the reasoning conversion problem of close-domain, we turned this problem into a language rewrite task for LLMs to be better at. We do not consider the handling of leaf nodes, as they are already the smallest unit of process description. We mainly consider the handling of non-leaf nodes. When dealing with non-leaf nodes, we know the operator type of the non-leaf node and the BPTS corresponding to all subtrees of the operator. Instead of attempting to comprehend the transformation rules by using the rich subject information and examples given to LLMs, we now take a more direct approach. By using input that resembles pseudocode to represent the relationship between an operator and the matching BPTS of its subtree, we may instruct LLMs to translate the input into more fluent plain language without altering its original meaning. Domain knowledge was eliminated as a result of the problem's transformation, and the instruction prompt was altered to \textbf{``Cover the input into fluent natural language without changing its meaning''}. The prompt templates for the four operators are shown in Figure \ref{fig:prompt_template}. In Algorithm \ref{rp}, the $constructPrompt$ function completes such construction.

\begin{algorithm}[!h]
    \SetKwData{node}{node}
    \SetKwData{children}{children}
    
    \SetKwInOut{Input}{input}
    \SetKwInOut{Output}{output} 
    
    \Input{maximum depth $DEPTH$, probability of generating zero child nodes $P_{ZERO}$, probability of generating two word nodes $P_{TWO}$, upper bound $NUM_{UP}$ and lower bound $NUM_{LOW}$ for generating other child nodes}
    \Output{Random tree} 
    \BlankLine 
    \emph{Initialize the queue $Q$ and add the root node $ROOT$ to the $Q$}\;
    \While {$Q$ is not empty \textbf{and} current depth is less than $DEPTH$}{
        \node $\leftarrow$ Take the first element of the queue $Q$\;
        \tcp{The probability of generating other number of nodes is 1-$P_{ZERO}$-$P_{TWO}$}
        \children $\leftarrow$ Construct child nodes based on $P_{ZERO}$, $P_{TWO}$, $NUM_{UP}$ and $NUM_{LOW}$\;
        \node.$children$ $\leftarrow$ \children\;
        Queue \children\;
    }
    \textbf{return} $ROOT$\;
    
    \caption{Random tree generation algorithm}
    \label{rtg} 
\end{algorithm}

\begin{algorithm}[!h]
    \SetKwData{node}{node}
    \SetKwData{child}{child}
    \SetKwFunction{RCPTGen}{RCPTGen}
    
    \SetKwInOut{Input}{input}
    \SetKwInOut{Output}{output} 
    
    \Input{Current node $N$}
    \Output{Rough Conditional Process Tree} 
    \BlankLine 
    \emph{This is a recursive algorithm called \RCPTGen}\;
    \eIf{The current node $N$ is a non-leaf node}{
        \eIf{The number of child nodes of $N$ is equal to 2}{
            Node $N$ is given as $\to$, $\times$, $\propto$ or $\wedge$\;
        }{
            Node $N$ is given as $\times$ or $\propto$\;
        }
        \If{Node $N$ is given as $\times$}{
            Add condition $c$ to $\times$\;
        }
        \For{\child : $N$.$children$}{
            \tcp{Recursively call yourself}
            \RCPTGen{child}\;
        }
    }{
        Node $N$ is given activity a\;
    }
    
    \caption{Rough Conditional Process Tree generation algorithm}
    \label{rcptg} 
\end{algorithm}

\begin{algorithm}[!h]
    \SetKwData{node}{node}
    \SetKwData{child}{child}
    
    \SetKwInOut{Input}{input}
    \SetKwInOut{Output}{output} 
    
    \Input{Rough Conditional Process Tree $R$}
    \Output{Conditional Process Tree} 
    \BlankLine 
    \emph{Initialize the queue $Q$, add the root node $ROOT$ of $R$ to the $Q$ and initialize flag $F$ to \textbf{FALSE}}\;
    \While {$Q$ is not empty}{
        \If{$F$ is \textbf{FALSE}}{
            \node $\leftarrow$ Take the first element of the queue $Q$\;
        }
        $F$ $\leftarrow$ \textbf{FALSE}\;
        \If{The current node \node is a non-leaf node}{
            \For{\child : \node.$children$}{
                \If{The operator of \node is the same as the operator of \child, and is $\to$ or $\wedge$}{
                    Remove the child node \child\;
                    $F$ $\leftarrow$ \textbf{TRUE}\;
                }
                \ElseIf{The operator of \node is $\propto$}{
                    Replace the first child node with condition c\;
                    $F$ $\leftarrow$ \textbf{TRUE}\;
                }
                \Else{
                    Queue \child\;
                }
            }
        }
    }
    
    \caption{Conditional Process Tree 
generation algorithm}
    \label{cpt1} 
\end{algorithm}

\subsection{Dataset Generation}
Due to CPT's structured nature, rules may be used to generate it in huge quantities, and when combined with our way to create process description text, the outcome is a large number of CPT-BPTS pairings that could potentially be PME domain datasets. The CPT is generated using the following rules: 1) Add a random number of child nodes to produce a random tree without node information. The precise method entails first defining the maximum $depth$, the probability of creating zero child nodes $p_{zero}$, the chance of generating two child nodes $p_{two}$, and the lower and upper bounds of other numbers of generating child nodes $num_{low}$ and $num_{up}$. Use breadth-first traversal to haphazardly add zero, two, or other numbers of child nodes to the current node. The depth of the created conditional process tree is less than or equal to depth since it will generate a zero number of child nodes. Refer to Algorithm \ref{rtg}. 2) According to the rules of CPT, nodes are assigned elements with a certain probability, which can be operators, conditions, and activities. The rule is as follows: a. If the number of sub-nodes of the current node is equal to 2, the operators $\to$, $\times$, $\wedge$ or $\propto$ can be assigned; b. If the number of child nodes of the current node is greater than 2, the operators $\to$ or $\wedge$ can be assigned; c. If the current node has no child nodes, assign an activity; refer to Algorithm \ref{rcptg}. 3) There are three irrational structures in the CPT that were produced using this rule: a. The child node is $\to$, and its parent node is also $\to$. b. The child node is $\wedge$, and its parent node is also $\wedge$. c. The current node is $\propto$, and its first child node is not a condition. Therefore, it is necessary to conduct a rationalization check on the CPT and remove or replace unreasonable subtrees, as shown in Algorithm \ref{cpt1}.

\section{Experiments}
\begin{figure}[htbp]
\centerline{\includegraphics[scale=0.34]{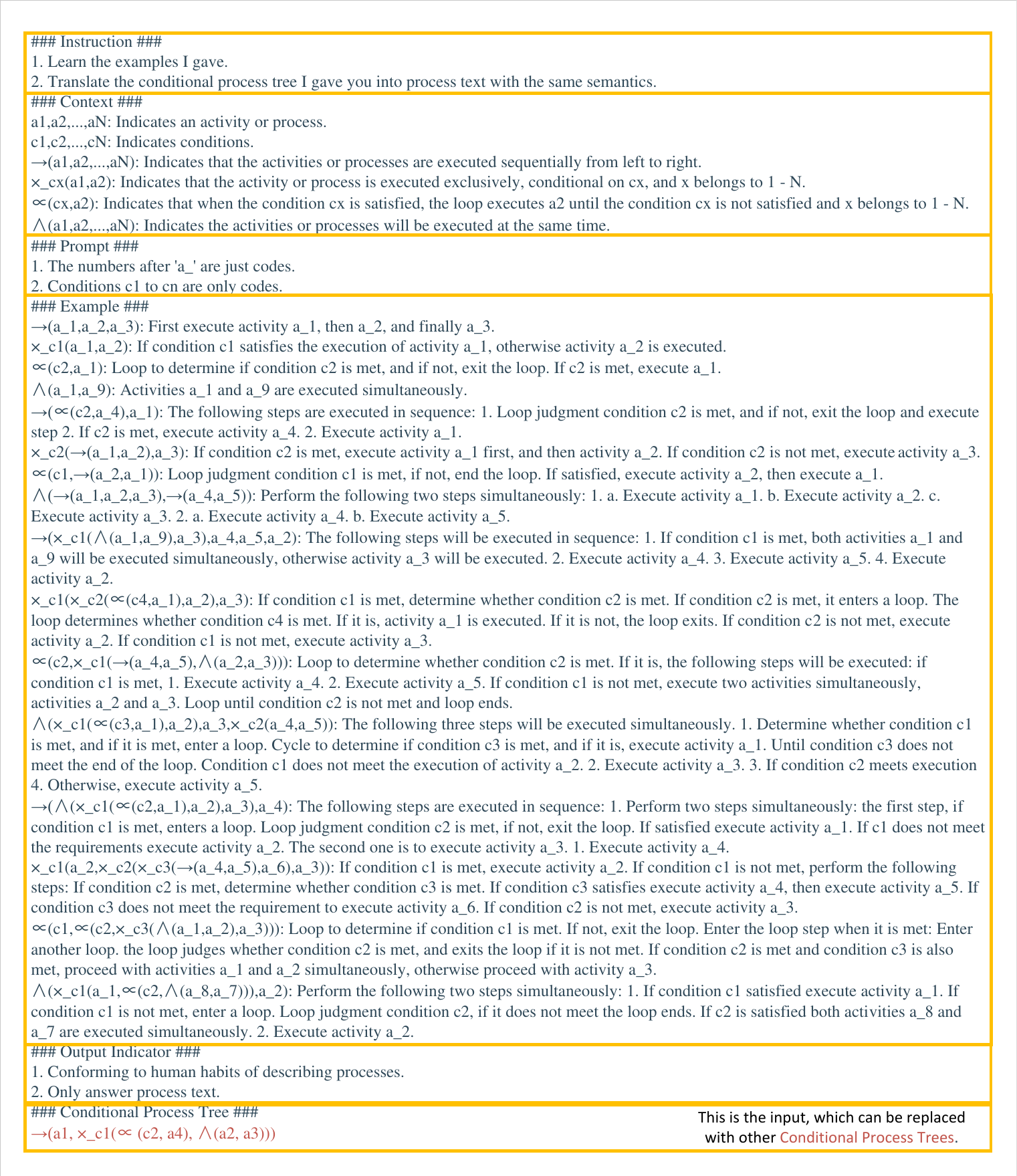}}
\caption{The traditional Few-shot In-Context Learning prompt method}
\label{fig:traditional}
\end{figure}

We contrasted our suggested approach with traditional prompts to show its efficacy, as shown in Figure \ref{fig:traditional}. The traditional prompt template consists of six components, whose names and purposes are shown below: 1) Instruction. The tasks that Large language Models (LLMs) are expected to complete. 2) Context. Introduction to background knowledge. 3) Prompt. Prevent LLMs from learning some toxic laws. 4) Example. Introduce Few-shot Learning. 5) Output Indicator. Specify the output format. 6) Conditional Process Tree. Input the Conditional Process Tree (CPT) sequence. We carefully selected CPTs of various difficulty levels. As shown in Table \ref{tab1}, the depth range of the conditional process tree is from 2 to 5, and every depth includes every operator. There are other CPTs that include many levels of loops and selections, with a maximum nesting level of 3. Finally, some CPTs with a number of nodes greater than 15 were selected. A total of 100 CPTs were selected.

\begin{table}[]
\centering
\caption{Composition of test dataset}
\label{tab1}
\begin{tabular}{@{}cll@{}}
\toprule
Aspect                      & Type            & Number       \\ \midrule
\multirow{4}{*}{Depth}      & = 2             & 10           \\
                            & = 3             & 20           \\
                            & = 4             & 20           \\
                            & = 5             & 20           \\
\multirow{2}{*}{Multilayer} & Loop            & 10           \\
                            & Selection       & 10           \\
Number of nodes             & \textgreater 15 & 10           \\
Total                       & Category 7      & Quantity 100 \\ \bottomrule
\end{tabular}
\end{table}

\begin{table}[htbp]
\centering
\caption{Test dataset statistics.}
\label{tab2}
\begin{tabular}{@{}cll@{}}
\toprule
Aspect                    & Type     & Number \\ \midrule
\multirow{2}{*}{Activity} & Max      & 20     \\
                          & Min      & 1      \\
\multirow{2}{*}{Node}     & Max      & 29     \\
                          & Min      & 3      \\
\multirow{2}{*}{Operator} & Max      & 12     \\
                          & Min      & 1      \\
Multi-layer Selection     & = 2 or 3 & 18     \\
Multi-layer Loop          & = 2 or 3 & 17     \\ \bottomrule
\end{tabular}
\end{table}

\begin{figure*}[htbp]
\centerline{\includegraphics[scale=0.43]{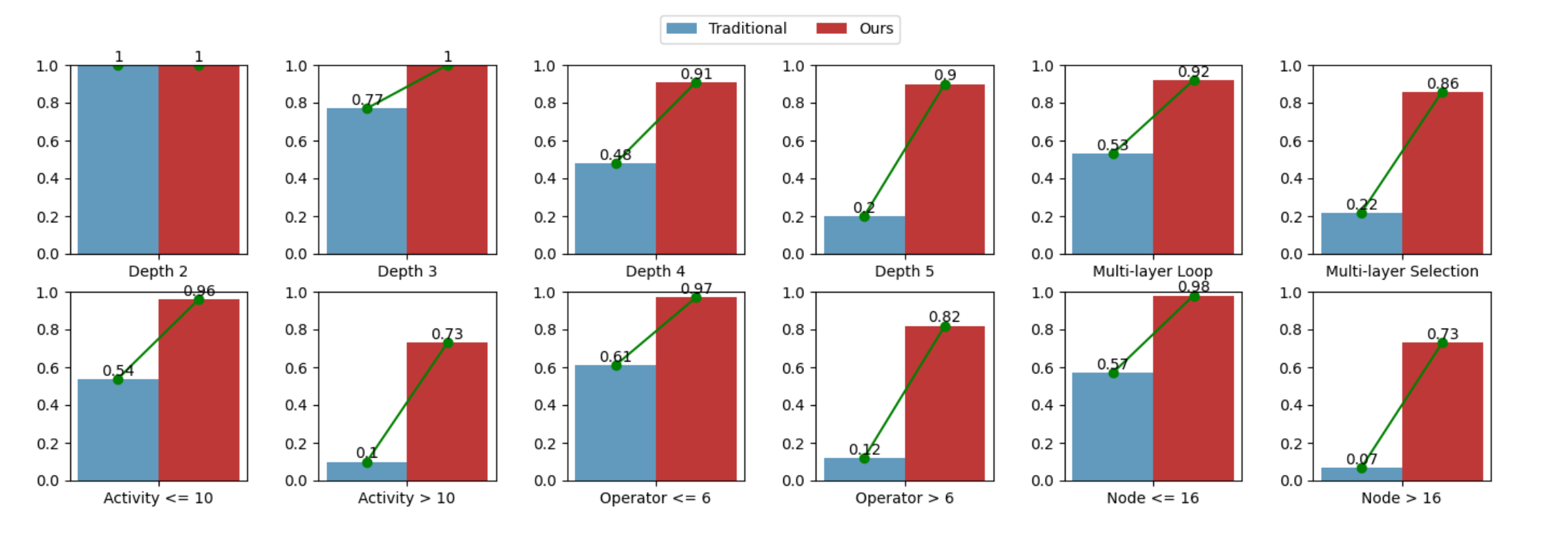}}
\caption{Comparison between traditional prompts and our method on 12 dimensions. We use the average of the maximum and minimum values of the number of activities, operators, and nodes in Table \ref{tab2} as the segmentation lines for simple and complex CPTs, respectively, to obtain the results in the second row of the figure.}
\label{fig:result}
\end{figure*}

The statistics of the selected CPTs are displayed in Table \ref{tab2}. 29 nodes are the maximum number, 20 activities are the maximum number, and 12 operators are the maximum number. For multilayer selection and looping, there are 18 and 17 CPTs, respectively.

The ``gpt-3.5 turbo'' model is what we use because it is the most potent model that OpenAI can access at the moment. To ensure the certainty of the answer generated by the model, we set the API parameter $temperature$ to 0, Set $top\_p$ to 1, and $n$ to 1. All other parameters have default values. Due to the lack of Source text, we enlisted the help of three graduate professionals to assess the effectiveness of creating Business Process Text Sketches (BPTSs). The evaluation criteria are: A score of 1 will be awarded if the generated BPTS completely matches the input CPT, there is no ambiguity, and the language is fluent. A score of 0 will be assigned if the generated BPTS and the input CPT do not agree. In the case of ambiguity in the generated BPTS, the score is based on the likelihood of correct understanding. For example, if there is an ambiguity, the likelihood of correct understanding is 0.5, so the score is 0.5. If there is complete correspondence, no ambiguity, but the language is not smooth, the evaluator's own feelings must be scored, and the score must be greater than 0.5.

\subsection{Overall performance}
We asked three evaluators to score the 100 data points generated using traditional prompts and the 100 data points generated using our approach according to the evaluation criteria. We took the average score of the three evaluators as the final score.

As shown in Table \ref{tab3}, our proposed method performs better overall than the traditional prompt method, with an overall accuracy improvement of 45.17\%. We compare the performance of the two methods from six aspects: the depth of the CPT, multi-layer loops, multi-layer selection, number of activities, number of operators, and number of nodes. In Table \ref{tab2}, we take the average of the maximum and minimum number of activities as the segmentation line for simple and complex CPTs. Similarly, the number of operators and nodes is divided in the same way. From Figure \ref{fig:result}, it can be seen that the accuracy of CPT conversion is 1.0 for the depth of 2. In addition, the performance of traditional prompts is lower than our method, and our method can greatly improve the accuracy of CPT conversion. In the conversion task of complex CPTs, our method has more obvious advantages, especially in the conversion of CPTs with a large number of nodes. The accuracy of our method is 10 times that of traditional prompts. Obviously, regardless of the method, as the complexity of CPT increases, its conversion accuracy will decrease.

\begin{table}[htbp]
\centering
\caption{Overall performance}
\label{tab3}
\begin{tabular}{@{}ll@{}}
\toprule
Method      & Accuracy \\ \midrule
Traditional & 48.25    \\
Ours        & \textbf{93.42*}   \\ \bottomrule
\end{tabular}
\end{table}

\subsection{Error analysis}
There are four main issues in the generated BPTS, namely: 1) loop condition errors. The correct BPTS should be ``execute the following activity cyclically if the condition is met'', but the generated BPTS should be ``execute the following activity cyclically if the condition is not met''. 2) Language confusion. When the input CPT has a cross-layer parallel structure, the parallel expression statements of the generated BPTS will become chaotic, making it difficult to understand the generated BPTS. 3) Parallel and sequential cannot be distinguished. In parallel statements, ``followed by'' will appear, while in sequential statements, ``at the same time'' will appear. 4) Hierarchy parsing error. When there are too many operators and the hierarchy is deep, this error is likely to occur. Some examples of errors are shown in Appendix \ref{A}.

\subsection{Case 1 business process document generation}
\begin{figure}[htbp]
\centerline{\includegraphics[scale=0.4]{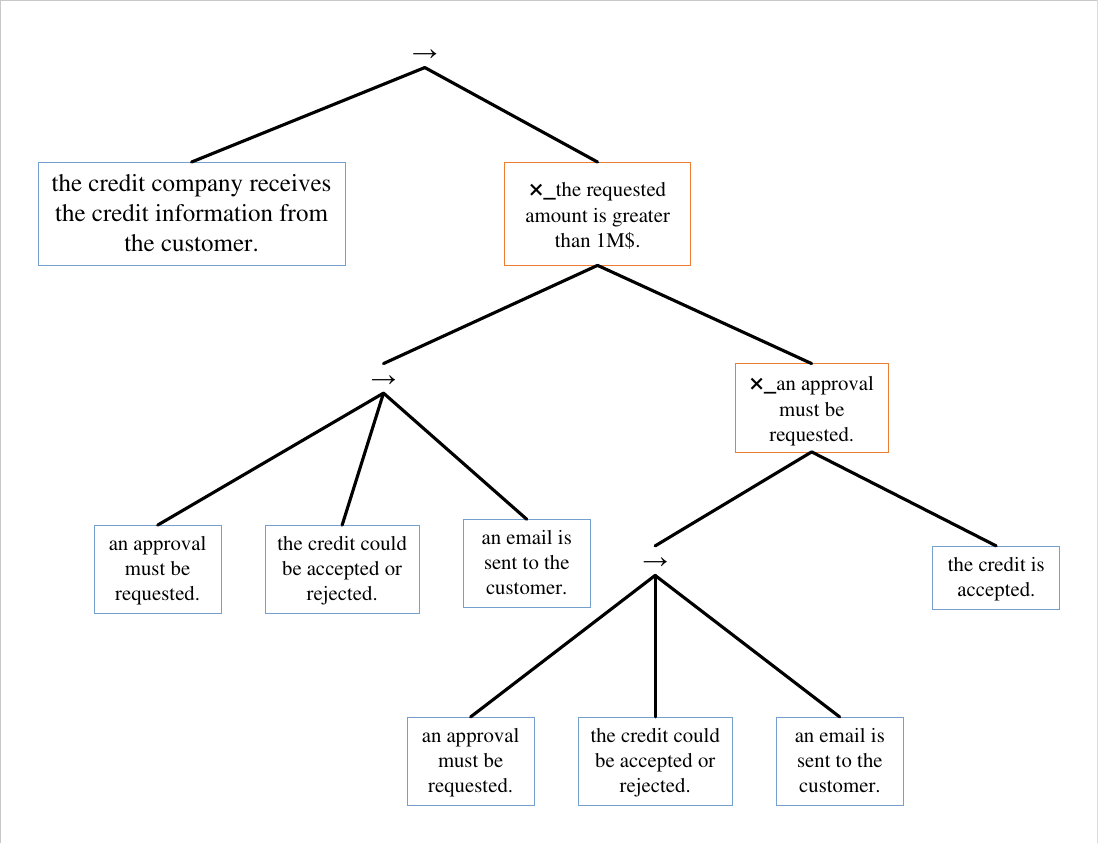}}
\caption{Instantiated CPT of bank loan. The blue box represents an activity, and the yellow box represents the conditions for exclusive selection.}
\label{fig:example}
\end{figure}

After the CPT is instantiated, it can be converted into business process description text through our method. We chose an instantiated CPT with a maximum number of layers not exceeding the experimental maximum, which mainly describes the business process of bank lending, as shown in Figure \ref{fig:example}.

\begin{figure}[htbp]
\centerline{\includegraphics[scale=0.565]{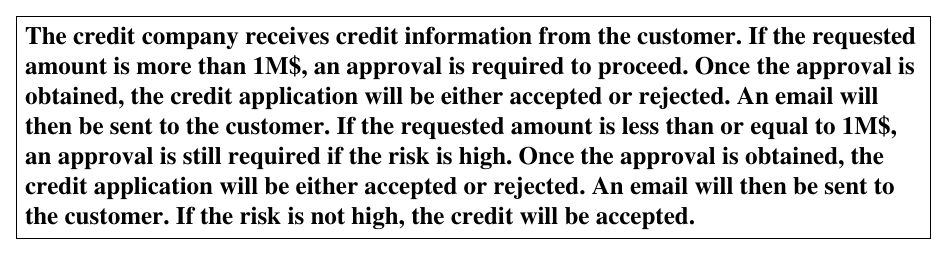}}
\caption{The result of the instantiated CPT}
\label{fig:example_result}
\end{figure}

The business process description text generated using our method is shown in Figure \ref{fig:example_result}. The generated usability process description text is unexpected; for example, the condition describes ``greater than'', and the opposite of the condition in the output BPTS is ``less than or equal'' rather than just ``less than''.

\subsection{Case 2 dataset generation}
Our algorithm can quickly generate a large number of CPTs. Among the generated 2500 CPTs, the maximum depth is 5, the maximum number of nodes is 40, the maximum number of activities is 25, and the maximum number of operators is 20. Generating 2500 CPTs with a maximum depth of 5 takes less than 2 seconds. We have selected three examples that are relatively difficult but have high-quality generated text. These three examples have different styles of descriptions, as shown in Appendix \ref{B}.

\section{Limitation}
The style of the BPTS generated by our method mainly depends on the style of the prompt template. Although the generated BPTS can accurately describe the CPT, it is closely related to the narrative style of the prompt template, and sometimes it may generate BPTS that do not conform to human descriptive habits. Furthermore, it is clear from the example of generating errors that there is a strong likelihood that the prompt is to blame for the error in generating BPTS. It is undeniable that the quality of prompt templates will directly affect the quality of generating BPTS, and we do not have a method to improve the quality of prompt templates. In addition, although our method generates BPTS with high accuracy, there is no method to detect the quality of the generated BPTS.

\section{Conclusion And Disscusion}
We propose a method for automatically generating Business Process Text Sketchs using a Large Language Model. Divide-and-conquer strategy-inspired, we broke down a complex CPT into numerous simple CPTs and solved each one separately. Then, we transformed the simple CPT inference transformation task into a language rewriting task that LLMs are better at. Experiments have shown that our method can achieve higher accuracy in generating BPTS compared to traditional prompt methods. Our work alleviates the awkward situation of lacking datasets in the domain of business process document generation, and our method also provides potential possibilities for dataset generation in the PME domain.

\vspace{12pt}
\bibliography{ref}

\begin{thebibliography}{10}

\bibitem{han2020bps}
Xue Han, Lianxue Hu, Lijun Mei, Yabin Dang, Shivali Agarwal, Xin Zhou, and
  Pengwei Hu.
\newblock A-bps: automatic business process discovery service using ordered
  neurons lstm.
\newblock In {\em 2020 IEEE International Conference on Web Services (ICWS)},
  pages 428--432. IEEE, 2020.

\bibitem{friedrich2011process}
Fabian Friedrich, Jan Mendling, and Frank Puhlmann.
\newblock Process model generation from natural language text.
\newblock In {\em Advanced Information Systems Engineering: 23rd International
  Conference, CAiSE 2011, London, UK, June 20-24, 2011. Proceedings 23}, pages
  482--496. Springer, 2011.

\bibitem{qian2020approach}
Chen Qian, Lijie Wen, Akhil Kumar, Leilei Lin, Li~Lin, Zan Zong, Shu’ang Li,
  and Jianmin Wang.
\newblock An approach for process model extraction by multi-grained text
  classification.
\newblock In {\em Advanced Information Systems Engineering: 32nd International
  Conference, CAiSE 2020, Grenoble, France, June 8--12, 2020, Proceedings 32},
  pages 268--282. Springer, 2020.

\bibitem{bellan2021process}
Patrizio Bellan, Mauro Dragoni, and Chiara Ghidini.
\newblock Process extraction from text: state of the art and challenges for the
  future.
\newblock {\em arXiv preprint arXiv:2110.03754}, 2021.

\bibitem{brown2020language}
Tom Brown, Benjamin Mann, Nick Ryder, Melanie Subbiah, Jared~D Kaplan, Prafulla
  Dhariwal, Arvind Neelakantan, Pranav Shyam, Girish Sastry, Amanda Askell,
  et~al.
\newblock Language models are few-shot learners.
\newblock {\em Advances in neural information processing systems},
  33:1877--1901, 2020.

\bibitem{openai2023gpt4}
OpenAI.
\newblock Gpt-4 technical report, 2023.

\bibitem{bubeck2023sparks}
S{\'e}bastien Bubeck, Varun Chandrasekaran, Ronen Eldan, Johannes Gehrke, Eric
  Horvitz, Ece Kamar, Peter Lee, Yin~Tat Lee, Yuanzhi Li, Scott Lundberg,
  et~al.
\newblock Sparks of artificial general intelligence: Early experiments with
  gpt-4.
\newblock {\em arXiv preprint arXiv:2303.12712}, 2023.

\bibitem{schick2020exploiting}
Timo Schick and Hinrich Sch{\"u}tze.
\newblock Exploiting cloze questions for few shot text classification and
  natural language inference.
\newblock {\em arXiv preprint arXiv:2001.07676}, 2020.

\bibitem{min2022rethinking}
Sewon Min, Xinxi Lyu, Ari Holtzman, Mikel Artetxe, Mike Lewis, Hannaneh
  Hajishirzi, and Luke Zettlemoyer.
\newblock Rethinking the role of demonstrations: What makes in-context learning
  work?
\newblock {\em arXiv preprint arXiv:2202.12837}, 2022.

\bibitem{lcl}
LI~Tong ZHU~Rui, LV~Changlong.
\newblock Automatic business process model deep generation based on ordered
  neurons long short term memory.
\newblock {\em Computer Integrated Manufacturing System}, 28(10):3225--3238,
  2022.

\bibitem{zhu2023tag}
Rui Zhu, Wenxin Li, and Canchang Jin.
\newblock Tag: Uml activity diagram deeply supervised generation from business
  textural specification.
\newblock In {\em 2023 IEEE International Conference on Software Analysis,
  Evolution and Reengineering (SANER)}, pages 956--961. IEEE, 2023.

\bibitem{bellan2022extracting}
Patrizio Bellan, Mauro Dragoni, and Chiara Ghidini.
\newblock Extracting business process entities and relations from text using
  pre-trained language models and in-context learning.
\newblock In {\em International Conference on Enterprise Design, Operations,
  and Computing}, pages 182--199. Springer, 2022.

\bibitem{zhou2022least}
Denny Zhou, Nathanael Sch{\"a}rli, Le~Hou, Jason Wei, Nathan Scales, Xuezhi
  Wang, Dale Schuurmans, Olivier Bousquet, Quoc Le, and Ed~Chi.
\newblock Least-to-most prompting enables complex reasoning in large language
  models.
\newblock {\em arXiv preprint arXiv:2205.10625}, 2022.

\bibitem{gao2023easy}
Chang Gao, Wenxuan Zhang, Wai Lam, and Lidong Bing.
\newblock Easy-to-hard learning for information extraction.
\newblock {\em arXiv preprint arXiv:2305.09193}, 2023.

\end{thebibliography}
\appendix
\section{Appendix}
\onecolumn
\subsection{Error generations}\label{A}
\begin{table*}[!h]
\centering
\caption{Example of error generation}
\label{tab4}
\begin{tabular}{@{}p{4cm}p{6cm}p{4cm}@{}}
\toprule
Conditional Process Tree                                                                                                                      & Business Process Text Sketch                                                                                                                                                                                                                                                                                                                                                                                                                                                                                                                                                                                                                                                                                                                                                                                                                                                                                                                                                                          & Reason for error                         \\ \midrule \\
$\propto(c1,\to(a\_4,\propto(c2,a\_6)))$                                                                                                                    & \begin{tabular}[c]{@{}p{6cm}@{}}Continuously check condition c1 until it is no longer true. If c1 is true, execute activity a\_4 and then continuously monitor condition c2 until it becomes false. Once c2 is false, perform activity a\_6.\end{tabular}                                                                                                                                                                                                                                                                                                                                                                                                                                                                                                                                                                                                                                                                                                                                                                                          & Loop condition error                     \\\\
\begin{tabular}[c]{@{}l@{}}$\to(a\_1,a\_2,\wedge(\to(a\_3,\wedge(a\_4,$\\$a\_5,a\_6,a\_7)),\to(\times\_c1(a\_8,$\\$a\_9),\propto(c2,a\_11),\wedge(a\_12$\\$,a\_13,a\_14,a\_15,a\_16),$\\$a\_17,a\_18),a\_19,a\_20,a\_21$\\$))$\end{tabular} & \begin{tabular}[c]{@{}p{6cm}@{}}The following steps will be carried out in a specific order. First, we will execute activity a\_1, followed by activity a\_2. Next, we will perform five steps concurrently. In the first step, we will execute activity a\_3, and then in the next four steps, we will execute activities a\_4, a\_5, a\_6, and a\_7 simultaneously. In the second step, we will execute activity a\_8 if condition c1 is met, otherwise, we will execute activity a\_9. We will then continuously check condition c2 until it is no longer true. Once it becomes true, we will run activity a\_11. After that, we will execute five steps simultaneously, which include activities a\_12, a\_13, a\_14, a\_15, and a\_16. In the fourth step, we will execute activity a\_17, and finally, in the fifth step, we will execute activity a\_18. In the third step, we will execute activity a\_19, and in the fourth step, we will execute activity a\_20. Lastly, in the fifth step, we will execute activity a\_21.\end{tabular} & Language confusion                    \\\\
\begin{tabular}[c]{@{}l@{}}$\to(a\_1,\propto(c4,a\_6),\wedge(\to(a\_7,$\\$a\_8,\times\_c2(a\_9,a\_10)),a\_11,$\\$a\_12,a\_13),\times\_c3(a\_14,$\\$a\_15))$\end{tabular}                                                     & \begin{tabular}[c]{@{}p{6cm}@{}}The following steps will be carried out in a specific order. First, activity a\_1 will be executed. Then, condition c4 will be continuously checked until it is no longer true. Once it becomes true, activity a\_6 will be run. At the same time, four steps will be carried out simultaneously. The first step involves executing activity a\_7, followed by activity a\_8. If condition c2 is met, activity a\_9 will be executed, but if it is not met, activity a\_10 will be executed instead. The second step involves executing activity a\_11, the third step involves executing activity a\_12, and the fourth step involves executing activity a\_13. Finally, if condition c3 is met, activity a\_14 will be executed. However, if condition c3 is not met, activity a\_15 will be executed instead.\end{tabular}                                                                                                                                                                                      & \begin{tabular}[c]{@{}p{3.5cm}@{}}Sequential relationships become parallel\end{tabular} \\\\ \bottomrule
\end{tabular}
\end{table*}
\onecolumn
\subsection{Successful generations}\label{B}
\begin{table*}[!h]
\centering
\caption{Three examples of successful generations with different styles}
\label{tab5}
\begin{tabular}{p{7cm}p{7cm}}
\toprule
Conditional Process Tree                                                                                                           & Process description text                                                                                                                                                                                                                                                                                                                                                                                                                                                                                                                                                                                                                                                                                                                                                                                                                                                                                                                                                                                                                                         \\ \midrule\\
\begin{tabular}[c]{@{}l@{}}
$\wedge(a\_1,\times\_c1(a\_2,\wedge(a\_3,\times\_c2(a\_4,a\_5))),a\_6,$\\$\propto(c4,\to(\wedge(a\_11,a\_12,a\_13),\propto(c5,a\_15),a\_16,$\\$\wedge(a\_17,a\_18,a\_19,a\_20,a\_21))))$\end{tabular} & \begin{tabular}[c]{@{}p{7cm}@{}}There are four steps that will be carried out at the same time. The first step is to execute activity a\_1. The second step involves checking if condition c1 is met. If it is, activity a\_2 will be executed. If not, two steps will be taken simultaneously. The first step is to execute activity a\_3, while the second step is to check if condition c2 is satisfied. If it is, activity a\_4 will be executed. If not, activity a\_5 will be executed instead. The third step is to execute activity a\_6. The fourth step involves continuously checking condition c4 until it is no longer satisfied. If it is, a specific process will be followed that involves multiple steps. Firstly, activities a\_11, a\_12, and a\_13 will be executed simultaneously. Secondly, condition c5 will be continuously checked until it is no longer true. Once it becomes true, activity a\_15 will be executed. Thirdly, activity a\_16 will be executed. Lastly, activities a\_17, a\_18, a\_19, a\_20, and a\_21 will be executed simultaneously in five steps.\end{tabular} \\\\
\begin{tabular}[c]{@{}l@{}}
$\to(\times\_c1(\times\_c2(a\_1,a\_2),a\_3),\wedge(a\_4,\times\_c3(a\_5$\\$,a\_6)),a\_7)$\end{tabular}                                                                       & \begin{tabular}[c]{@{}p{7cm}@{}}The following steps will be carried out in a specific order. First, if condition c1 is met, activity a\_1 will be executed only if condition c2 is also met. However, if condition c2 is not met, then activity a\_2 will be executed instead. If condition c1 is not fulfilled, then activity a\_3 will be executed.\\ Secondly, two steps will be carried out simultaneously. The first step involves executing activity a\_4. The second step involves checking if condition c3 is met. If it is met, activity a\_5 will be executed. If it is not met, activity a\_6 will be executed instead.\\ Lastly, activity a\_7 will be executed.\end{tabular}                                                                                                                                                                                                                                                                                                                                                                       \\\\
\begin{tabular}[c]{@{}l@{}}$\to(\propto(c3,\propto(c4,a\_7)),a\_8,\wedge(\propto(c5,a\_10),\to($\\$\times\_c2(a\_11,a\_12),\wedge(a\_13,a\_14,a\_15,a\_16,a\_17$\\$))))$\end{tabular}                                     & \begin{tabular}[c]{@{}p{7cm}@{}}The process will follow these steps in order:\\ 1. Check condition c3 continuously until it is no longer true. Once it is met, check condition c4 until it is no longer true. When c4 becomes true, execute activity a\_7. 2. Execute activity a\_8. 3. Two steps will occur simultaneously. The first step involves continuously checking condition c5 until it is no longer true. Once c5 becomes true, run activity a\_10. The second step will proceed as follows: if condition c2 is met, perform activity a\_11. If not, perform activity a\_12 instead. After that, execute activities a\_13, a\_14, a\_15, a\_16, and a\_17 simultaneously.\end{tabular}                                                                                                                                                                                                                                                                                                                                                                        \\\\ \bottomrule
\end{tabular}
\end{table*}
\end{document}